\def\year{2019}\relax
\documentclass[letterpaper]{article} 
\usepackage{aaai19}  
\usepackage{times}  
\usepackage{helvet} 
\usepackage{courier}  
\usepackage[hyphens]{url}  
\usepackage{graphicx} 
\usepackage{graphicx}  
\usepackage{algorithm,algpseudocode}
\usepackage{amsmath}
\usepackage{graphicx}
\usepackage{textcomp}
\usepackage{xcolor}
\usepackage{mathtools}
\usepackage{multirow}
\usepackage{url}
\usepackage{hyperref}
\usepackage{mathtools}
\usepackage{amsmath,amssymb,amsfonts}
\usepackage{booktabs}
\usepackage[normalem]{ulem}
\usepackage{graphicx}
\usepackage{pdfpages}
\usepackage{cancel}
\usepackage{varioref}
\usepackage{booktabs}
\usepackage{commath}

\DeclarePairedDelimiter\ceil{\lceil}{\rceil}

\title{Towards Semantic Noise Cleansing of Categorical Data\\ based on Semantic Infusion}
\author{ \Large \textbf{Rishabh Gupta \and Rajesh N Rao}\\ 
Research and Technology Center, Robert Bosch, Bengaluru, India\\
\{Gupta.Rishabh, RajeshNagaraja.Rao\}@in.bosch.com 
}
 
\begin{document}

\maketitle

\begin{abstract}
Semantic Noise affects text analytics activities for the domain-specific industries significantly. It impedes the text understanding which holds prime importance in the critical decision making tasks. In this work, we formalize semantic noise as a sequence of terms that do not contribute to the narrative of the text. We look beyond the notion of standard statistically-based stop words and consider the semantics of terms to exclude the semantic noise. We present a novel Semantic Infusion technique to associate meta-data with the categorical corpus text and demonstrate its near-lossless nature.  Based on this technique, we propose an unsupervised text-preprocessing framework to filter the semantic noise using the context of the terms. Later we present the evaluation results of the proposed framework using a web forum dataset from the automobile-domain. 
\end{abstract}

\section{Introduction}
\label{sec:intro}
The structured text has become ubiquitous in recent years owing to increased digitization and automation in several industries. The structured domain-specific text presents a unique opportunity in aiding decision-making processes and activities.  The traditional text analytics techniques~\cite{aggarwal2012mining} are the first choice to perform these extraction and mining tasks, but in a domain-specific setup, they are challenged by \emph{Semantic Noise}. In the field of Communication Systems~\cite{brogan1974_semantic_noise}, semantic noise denotes a type of disturbance in the transmission of a message that interferes with its interpretation. In the domain of structured text, we define semantic noise as a sequence of terms (sentences) that do not contribute to the narrative of the text. In fact, these terms may give completely orthogonal information to the true narrative of the text.
For e.g., consider the following complaints as registered by consumers regarding the problems in the automobile domain: \emph{``Rear latch/striker failed in accident. Colorado state police.''} and \emph{``When applying brakes, excessive effort is necessary to try to top. System replaced several times. Please describe details''}.  In these complaints, the sentences \emph{``Colorado state police.''} and \emph{``Please describe details''} can be marked as semantic noise because they do not contribute to the understanding of the automobile problems. Filtering them can help a domain expert in the effective resolution of the problems. In addition, we also define the sentences as semantic noise, which are non-relevant such as ``opinions’’ or cross-topic such as ``political’’ in the automobile domain.

In literature, researchers have proposed several systems to remove the semantic noise that depends on either a predefined list of domain-specific stop words ~\cite{Baradad2015CorpusSS}  or a computation mechanism that generates this list~\cite{Ayral2011AnAD}~\cite{Lo2005AutomaticallyBA}.  In practice, these systems have many limitations such as the predefined list becomes outdated very quickly and thus requires regular revisions and updates~\cite{sinka2003evolving}. In addition, the computation mechanisms to generate the domain-specific stop words are largely based on document frequency filtering or term frequency schemes~\cite{forman2004pitfall}. These mechanisms fail especially for the cases when the corpus is discordantly distributed across the categories of the domain. Recently, researchers have proposed outlier detection systems while creating semantic clusters in the vector space~\cite{camacho2016find}. In practice, for semantic noise removal tasks, these systems require manual intervention to identify the relevant clusters, as they do not have a notion of an association between the meta-data and semantic clusters. 




We present a novel \emph{Semantic Infusion} technique which helps in associating meta-data with the corpus text when represented in a vector space. In this technique, we infuse special markers (referred to as Anchors) within sentences of the corpus. Using the infused corpus, we then can obtain the relevant semantic clusters within the neighborhood of anchors in the vector space without any manual intervention. 
\begin{figure*}[htbp]
	\includegraphics[width=\linewidth]{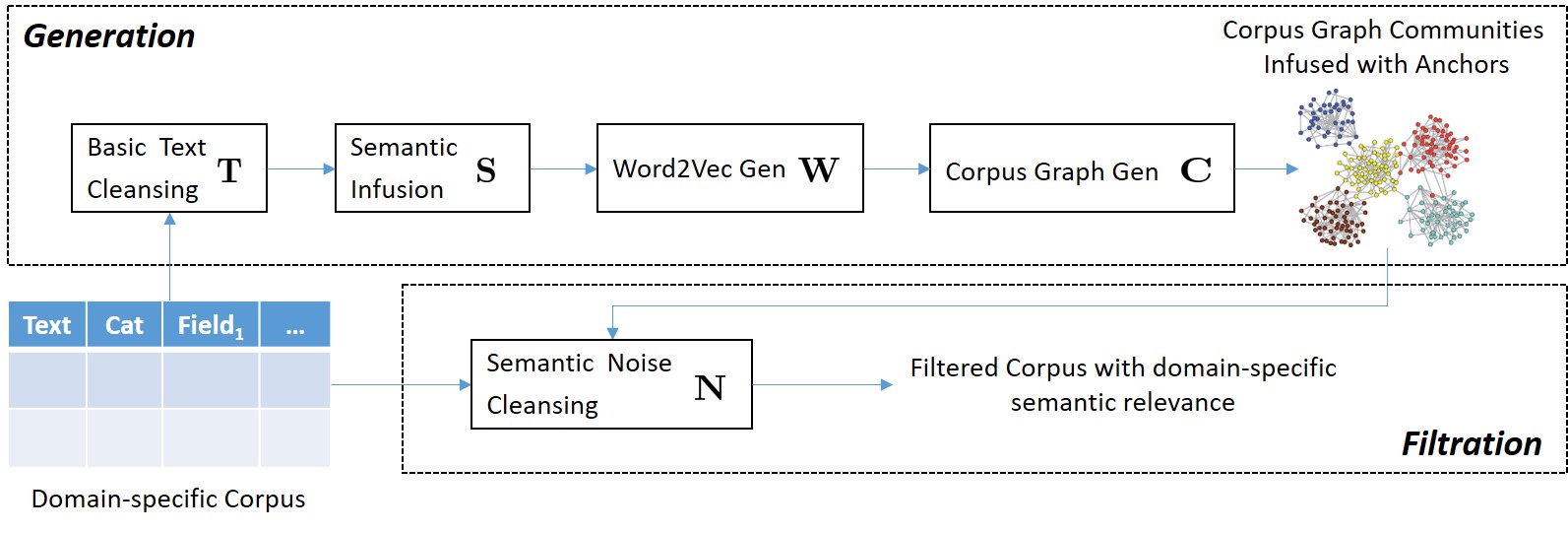}
	
	\caption{The proposed text-preprocessing framework which consists of $5$ individual modules: Basic Text Cleansing ($\mathbf{T}$), Semantic Infusion ($\mathbf{S}$), Word2Vec Gen ($\mathbf{W}$), Corpus Graph Gen ($\mathbf{C}$) and Semantic Noise Cleansing ($\mathbf{N}$).}
	\label{fig:dipsi_framework}
\end{figure*}
To showcase this, we present a text-preprocessing framework to filter out the semantically noisy sentences in a categorical corpus. We choose an automobile domain web forum dataset known as  NHTSA (National Highway Traffic Safety Administration)~\cite{NHTSA} and present a  study. In this, we motivate the usefulness of semantic infusion and demonstrate its near-lossless nature based on the Pairwise Inner Product (PIP) loss metric~\cite{yin2018dimensionality}. Later, we report our observations and evaluation results of the proposed framework based on our experiments.

 

\section{The Proposed Framework}
\label{sec:dipsi_framework}
In this section, we describe the proposed text-preprocessing framework which consists of $5$ individual modules as shown in Fig.~\ref{fig:dipsi_framework}. A domain-specific categorical corpus of $N$ documents $\mathcal{D}=\{d_k\}_{k=1}^{N}$ is given as input to the framework. Within this corpus, each document $d_k$ contains a set of sentences $j_{d_k}$ and belongs to a class $c_i$ in the set of $M$ classes  $\mathcal{C}=\{c_i\}_{i=1}^{M}$. For each document $d_k$, the framework finds a set of sentences $p_{d_k}$ ($p_{d_k} \subset j_{d_k}$) which can be treated as semantic noise. The detailed explanation of each module of the framework is as follows:

\subsection{Basic Text Cleansing ($\mathbf{T}$)}
\label{sub-sec:b_text_clean}
For each document $d_k$ in the corpus $\mathcal{D}=\{d_k\}_{k=1}^{N}$, this module performs the basic cleaning of the document's sentences such as removal of symbols and special characters. Also, this module removes the language-specific stop words e.g., for the English language it removes the common stop words such as ``his'', ``and'', ``he'', ``the'' etc. The list of stop words for a specific language can be easily obtained using various resources available on-line~\cite{stop-words}.

\subsection{Semantic Infusion ($\mathbf{S}$)}
\label{sub-sec:sem_inf}
This module takes the clean sentences as generated by the previous module $\mathbf{T}$ and performs the semantic infusion technique. The basic idea of this is to infuse additional meta-data (referred to as \textbf{Anchors}) within the clean sentences so that the vector space (as generated by Word2Vec Gen ($\mathbf{W}$) module) can be partitioned into the labeled regions. Intuitively, this technique helps in the automatic detection of relevant semantic clusters by correlating the position of anchors with the likelihood of their co-occurrence with the semantically significant terms. Later, this enables the Corpus Graph Gen ($\mathbf{C}$) module to select the contextually relevant communities in an unsupervised fashion. 

Given a clean sentence of length = $len$, of a document $d_k$ and class $c_i$, the semantic infusion technique defines the \emph{Infusion Frequency ($I_{freq}$)}, where $I_{freq} \in \mathbb{R}$, as the count of anchors to be infused in the clean sentence. The $I_{freq}$ is computed as given by the Equation~\ref{eqn:Infusion_freq}, where the logarithmic function ensures that the $I_{freq}$ $\cancel{\propto}$ $len$. This helps in making this technique a near-lossless in nature as demonstrated further in the Experiments section below.  
\begin{equation}
\label{eqn:Infusion_freq}
I_{freq} = \ceil*{\frac{\log_{2}{(len)}}{2}}
\end{equation}

Next, the semantic infusion technique generates non-consecutive random numbers whose count = $I_{freq}$ and the range =  $[0, len -1 ]$. Considering each random number as an index, the technique infuses an anchor = ``$A\_c_i$" in the clean sentence to get the final infused sentence. 

E.g., a clean sentence \emph{``right front wheel locked vehicle spin response anti lock brakes''} of a document class $c_i$ = \emph{Service-Brakes} will be processed by this module as  \emph{``right A\_Service-Brakes front wheel locked vehicle spin A\_Service-Brakes response A\_Service-Brakes anti lock brakes''}.

\subsection{Word2Vec Gen ($\mathbf{W}$)}
\label{sub-sec:w2v}
This module takes the infused sentences as generated by the previous module $\mathbf{S}$ and further generates the word vectors. These vectors capture the co-occurrence statistics of the words, such that, words that typically co-occur or words that share similar context are closer to each other in the vector space. These vectors are generated by an unsupervised algorithm named known as Word2Vec~\cite{mikolov2013efficient}.

\subsection{Corpus Graph Gen ($\mathbf{C}$)}
\label{sub-sec:corpus_graph}
This module takes the word vectors as generated by the previous module $\mathbf{W}$ and executes the $3$ step corpus graph generation algorithm. The intention of Corpus Graph is to identify the most relevant semantic clusters i.e., the set of words that typically represent a single context for each document class $c_i$. This in turn, helps in removing the semantically noisy words from the class $c_i$. The $3$ steps of the module $\mathbf{C}$ are explained as follows:\\

\noindent($1$) \textbf{Graph Building:} 
This step builds a weighted graph, $G$ = ($V$,$E$), using the words (as nodes) in the dataset and their word vectors as generated by the module $\mathbf{W}$. Given two words $a$ \& $b$ in the dataset and their word vectors $V_a$ \& $V_b$, the distance ($D_{ab}$) between them is defined using the cosine similarity metric \cite{salton1988term} as given by the Equation \ref{eqn_cosine_sim}.

\begin{equation}
\label{eqn_cosine_sim}
D_{ab}= \frac{V_a \cdot V_b}{\parallel V_a\parallel \parallel V_b\parallel}
\end{equation}

An edge $E_{ab}$ is drawn in the graph $G$, between two nodes $a$ and $b$, if
the $D_{ab}$ between them is greater than the threshold value $\theta$ i.e., $D_
{ab} > \theta$. Based on our experiments with various values of $\theta$, we consider $\theta=0.6$.
In addition, an edge weight $W_{E_{ab}}$ is assigned 
to each edge $E_{ab} \in E$, as per equation \ref{eqn:edge_weight}.

\begin{equation}
\label{eqn:edge_weight}
W_{E_{ab}} = \frac{1}{1 - D_{ab}}
\end{equation}

Intuitively, this means that only the similar contextual words (nodes) are 
connected (with an edge) in the graph $G$, where edge weight being the measure 
of their similarity.\\

\noindent($2$) \textbf{Graph Clustering:} 
This step identifies the graph communities $C = \{C_1 \dots C_M\}$ where $M \in 
\mathbb{R}$, within the weighted graph $G$ generated in the Step $(1)$. 
Intuitively, each graph community contains the \emph{Semantic Concepts} 
present in the dataset i.e., the set of words that typically represent a single
context. These concepts need not be precisely be entities or relations, but can be words that typically co-occur in a single context. For example, in document class $c_i = $ \emph{Seat-Belts}, concepts (words) associated with it: \emph{`belt', `retractor', `passive', `restraint', `retracted', `retract', `belts', `seatbelt', `lap', `fasten', `retain', `motorized', `unbuckle',} etc. form a semantic context and typically co-occur in single community.

The communities are detected using the \emph{Parallel Louvian Method (PLM)} \cite{blondel2008fast}  graph clustering algorithm in a recursive fashion. Detecting the clusters in a graph is an $NP-hard$ problem, and thus \emph{PLM} algorithm applies heuristics to find a locally optimal solution. 
In addition, the PLM algorithm can be parallelized which makes it extremely fast  to run on a large corpus~\cite{staudt2016engineering}~\cite{emmons2016analysis}. 
The stopping criteria for the recursion is defined using the \emph{Modularity Index} \cite{newman_modularity} as given by Equation \ref{eqn:modularity_index}.

\begin{equation}\label{eqn:modularity_index}
Q = \frac{1}{2m} \sum_{a,b} \left[ W_{E_{ab}}  - \frac{k_a k_b}{2m}\right] \delta(c_a, c_b)
\end{equation}

where $W_{E_{ab}}$ is the edge weight between nodes $a$ and $b$, $k_a =
\sum_b{W_{E_{ab}}}$, $c_a$ is the community to which $a$ belongs. Here, $\delta(x,
y) = 1$ if $x = y$, $\delta(x, y) = 0$ if $x \ne y$ and $m =
\frac{1}{2}\sum_{ab}W_{E_{ab}}$.\\

The recursive graph clustering algorithm provides a hierarchical representation
of the communities in which the higher level community represents semantic details like
sub-domains and lower level communities represents semantic concepts as shown in Table \ref{tab:community_sample} below. Note, in this work we consider only the third level (in the hierarchy) communities that have a minimum membership of $k=3$ nodes (words) and the rest of the communities are discarded as noise.\\

\noindent($3$) \textbf{Graph Selection:} This step selects a set of \emph{Anchored Communities} $C_A$ from the graph communities $C = \{C_1 \dots C_M\}$, as identified in the Step $(2)$, such that $C_A \subset C$ and $C_A = \{c_1 \dots c_N\}$ where $N \in \mathbb{R}$ and $N < M$. The $C_A$ communities are the ones which have at-least one \emph{Anchor} as a part of their semantic concepts. For e.g., in Table \ref{tab:community_sample}, $1-24-3$ and $1-24-4$ are the anchored communities as they have \emph{Anchor}: \emph{A\_Equipment}\ and \emph{A\_Fuel-System} respectively.

The basic idea of this step is to remove the non-anchored communities from the set of communities $C$. This is because the non-anchored communities might contain either the words which do not represent any semantic concepts or the language specific words which can be treated as the semantic noise.

\subsection{Semantic Noise Cleansing ($\mathbf{N}$)}
\label{sub:sem_noise_rem}
This module leverages the anchored communities $C_A = \{c_1 \dots c_N\}$ as selected by Step $(3)$ of the module $\mathbf{C}$ and identifies the sentences in each document $d_k$ which can be treated as semantic noise. The sentences are marked as noise 
based on the value of the \emph{Community Encoded Vector} which is defined as follows:

For the entire set of \emph{Anchored Communities} $C_A$, in which for each community $c_i$ the semantic concepts are given as $W(c_i) = {w^i_1, w^i_2, ... , w^i_{|c_i|}}$ and the community encoded vector $V_C$ of length $N$ is given as $V_C = \{vc_1, vc_2, ..., vc_N\}$ where the $i^{th}$ element of $V_C$ is represented by $vc_i$ i.e., each element represent an anchored community. The vector $V_C$ is initialized with all its element equals to $0$. 
%
Now for each sentence consisting the sequence of terms $t_1$, $t_2$, $\dots$, $t_K$, the vector $V_C$ is updated sequentially as per the Equation \ref{eqn:comm_vector_update}, where $f(vc_i) = 1$ is the increment function.
A sentence is marked as semantic noise if norm of the community encoded
vector for the sentence is zero, i.e., $||V_C|| = 0$.


\begin{equation}\label{eqn:comm_vector_update}
vc_i = \begin{cases}
vc_i + f(vc_i) & \text{if $t_i \in W(c_i)$ } \\
vc_i & \text{otherwise}
\end{cases}
\end{equation}

\begin{table*}[h]
	\caption{A Sample Consumer Complaint registered on the NHTSA platform.}
	
	\label{tab:sample_complaint}
	\begin{tabular}{l|c}
		\toprule
		Ticket Id   & 705071                                                                                                                                                                                                                                                                                                                                                                                                                                                                                                                                                                \\
		\midrule
		Company     & Ford                                                                                                                                                                                                                                                                                                                                                                                                                                                                                                                                                                 \\
		\midrule
		Model       & Excursion                                                                                                                                                                                                                                                                                                                                                                                                                                                                                                                                                             \\
		\midrule
		Make        & 2001                                                                                                                                                                                                                                                                                                                                                                                                                                                                                                                                                                  \\
		\midrule
		Date        & 20060516                                                                                                                                                                                                                                                                                                                                                                                                                                                                                                                                                              \\
		\midrule
		Component   & Seat-Belts                                                                                                                                                                                                                                                                                                                                                                                                                                                                                                                                                            \\
		\midrule
		Ticket Text & \begin{tabular}[r]{@{}c@{}}I have a 2001 excursion and the driver's side seat belt pops free by it's self several times during a couple of\\ hours of driving. Dealer said it wasn't covered. I also had the windows \& door locks stop working  back in\\ 2006. It needed a new gem module and some fuse panel work (due to a leak in the windshield  seal from the\\ factory). I finally saved up enough money to get everything fixed-- roughly \$1100.00. The rep. From the window\\ company said he has seen several like this himself.\end{tabular}\\
		\bottomrule
		
	\end{tabular}
\end{table*}

\section{Experiments}
\label{sec:exp}
In this section, we describe the categorical dataset: NHTSA and our experimental testbed using which we demonstrate the near-lossless nature of the semantic infusion technique based on the Pairwise Inner Product (PIP) loss metric~\cite{yin2018dimensionality}. In addition, we also present a manual evaluation mechanism using which we compute the performance results of the proposed text-preprocessing framework in filtering the semantically noisy sentences from the corpus.

\subsection{Dataset}
\label{sub-sec:dataset}
In this work, we take complaints of the automobile domain which are registered by the consumers at a web forum platform known as NHTSA (National Highway Traffic Safety Administration)~\cite{NHTSA}. A sample consumer complaint on NHTSA platform is shown in the Table~\ref{tab:sample_complaint} and statistics of dataset is given in the Table~\ref{tab:data_stats}. We extract following $2$ columns from this dataset: ``Component" and ``Ticket Text".  
%
%
%


\vspace{-0.4cm}
\begin{table}[H]
	\centering
	\caption{The statistics of NHTSA dataset.}
	\vspace{0.1cm}
	\label{tab:data_stats}
	\begin{tabular}{l|c}
		\toprule
		\#  Component  & 21                                                                                                                                                                                                                                                                          \\
		
		\midrule
		\# Consumer Complaints       & 70,000 (14 x 5,000)                                                                                                                                                                                                                                                        \\
		\midrule
		Component Classes    & \begin{tabular}[c]{@{}c@{}} Seat-Belts, Child-Seat,\\ Seats, Wheels, Tires,\\  Latches-Locks-Linkages,\\ Equipment,   Service-Brakes,\\ Electronic-Stability-Control,\\     Fuel-Propulsion-System,\\  Visibility-Wiper, Fuel-System,  \\ Visibility,  Exterior-Lighting  \end{tabular} \\
		
		\midrule
		\begin{tabular}[c]{@{}l@{}}Columns Used\\ in this Work \end{tabular}  & \begin{tabular}[c]{@{}l@{}}Component \\ Ticket Text\end{tabular}   \\
		\bottomrule
	\end{tabular}
\end{table}

\subsection{Generation}
We use the extracted NHTSA dataset as input to the proposed framework. In turn, it returns the communities (third level) $C = \{C_1 \dots C_M\}$, where $M = 147$, and anchored communities $C_A = \{c_1 \dots c_N\}$, where $N = 11$, using the following $4$ modules: Basic Text Cleansing ($\mathbf{T}$), Semantic Infusion ($\mathbf{S}$), Word2Vec Gen ($\mathbf{W}$) and Corpus Graph Gen ($\mathbf{C}$). These results are summarized in the Table \ref{tab:community_gen}. A snapshot of the generated communities is shown in Table~\ref{tab:community_sample}, where numbers in the left $1-24-x$ indicate the location of community in the hierarchy of all communities. In the snapshot, we are considering the $24^{th}$ community in second level hierarchy in which each third level community captures a specific semantics context.  The snapshot also shows the third level anchored communities: $1-24-3$ and $1-24-4$ with \emph{Anchor}: \emph{A\_Equipment}  and \emph{A\_Fuel-System} \& \emph{A\_ Fuel-Propulsion-System} respectively.\\

\noindent \textbf{Observation:}  There are $3$ component class pairs: Service-Brakes \& Electronic-Stability-Control, Fuel-System \& Fuel-Propulsion-System, and Visibility \& Visibility-Wiper which have shared anchored community (as mentioned in the Table~\ref{tab:community_gen}). This is due to the use of common words to explain the problems in these component class pairs. 
\vspace{-0.4cm}
\begin{table}[H]
	\centering
	\caption{The statistics after Corpus Graph Gen ($\mathbf{C}$) module.}
	\vspace{0.1cm}
	\label{tab:community_gen}
	\begin{tabular}{l|c}
		\toprule
		\begin{tabular}[c]{@{}c@{}}\# 3rd Level Communities\\ Generated (M)\end{tabular}                                 &  147                                                                                                                                                                                                                                                                                     \\
		\midrule
		\begin{tabular}[c]{@{}c@{}}\# Anchored Communities\\ Selected (N)\end{tabular}                                   & 11                                                                                                                                                                                                                                                                                    \\
		\midrule
		\begin{tabular}[c]{@{}c@{}}Component Classes with\\ shared\\ Anchored Communities\end{tabular}           & \begin{tabular}[c]{@{}c@{}} - Service-Brakes \&  \\  Electronic-Stability-Control\\ - Fuel-System \&  \\    Fuel-Propulsion-System\\ - Visibility \& Visibility-Wiper\end{tabular}\\
		\bottomrule                                                                                               
	\end{tabular}
\end{table}

%

%
%
%
%
%
%
%
\vspace{-0.5cm}
\begin{table}[h]
	\centering
	\caption{A snapshot of the generated communities.}
	\label{tab:community_sample}
	
	\begin{tabular}{|c|l|}
		\toprule
		\centering 1-24-3 & \begin{tabular}[c]{@{}l@{}}{[}`blowing', `cold', `heater', `temperature', `hot',\\ `defroster', \textbf{`A\_Equipment'}, `conditioner', \\ `defogger', `fan', `defrost', `degrees', `temp', \\ `setting', `blower', `heat', `conditioning' \\
			
			 
			  `cool', `condenser', `hvac', `summer' ,`blows' {]}\end{tabular} \\
		\midrule
		\centering 1-24-4 & \begin{tabular}[c]{@{}l@{}}{[}`solenoid', `pump', \textbf{`A\_Fuel-System'}, `valve', \\ `oxygen', `line', `fuel', `injectors', `tank', \\
			
			 
			  `egr', `injector', `carburetor', `pipe', `injection', \\ `tanks', `filter', `vacuum', `sending', `hoses', 
			  
			  
			  \\ `pumps', `inlet', `supply', `mass', `clogged', \\ `port',  `units',\textbf{`A\_Fuel-Propulsion-System'}{]}\end{tabular}               \\
		\bottomrule         
	\end{tabular}
\end{table}
\begin{figure*}[htbp]
	\centering
	\begin{equation}\label{eqn:bias_var}
	\mathbb{E}[\norm{{E}{E^T} - {\hat{E}}{\hat{E}_T}}] \approxeq \sqrt{\sum\limits_{i = k+1}^{d}\lambda_i^{4\alpha} } + 2\sqrt{2n}\alpha\sigma\sqrt{\sum\limits_{i =1}^k\lambda_i^{4\alpha-2}}+\sqrt{2}\sum\limits_{i=1}^k(\lambda_i^{2\alpha} - \lambda_{i+1}^{2\alpha})\sigma\sqrt{\sum\limits_{r\leq i < s}(\lambda_r - \lambda_s)^{-2}}
	\end{equation}
\end{figure*}
\vspace{-0.2cm}
\subsection{Semantic Infusion is Near-Lossless in Nature}
Given that, the word embeddings capture the word relations of a given corpus and the dimensionality of embeddings represents the quality of these relations~\cite{yin2018dimensionality}. We demonstrate the near-lossless nature of semantic infusion technique while studying the change in the dimensionality of the word embeddings (hence the quality of word relations), from basic to the infused corpus as obtained after the Basic Text Cleansing ($\mathbf{T}$) and Semantic Infusion ($\mathbf{S}$) modules respectively. We leverage the earlier work~\cite{yin2018dimensionality}, which states that the optimal dimensionality ($k^\ast$) of the word embeddings for a particular corpus, as the one which minimizes the Pairwise Inner Product (PIP) loss as given by the Equation \ref{eqn:bias_var}. In this, $E = U.,_{1:d}D^{\alpha}_{1:d,1:d}$ is the oracle embedding and $\hat{E} = \hat{U}.,_{1:d}\hat{D}^{\alpha}_{1:k,1:k}$ is the trained embedding, consisting of signal directions ($U$) and their magnitudes ($D^\alpha$), symmetric with spectrum ${\lambda}^d_{i=1}$, for any $0\leq\alpha\leq1$ and $k \leq d$, symmetric with zero mean, variance $\sigma^2$ entries.
\vspace{-0.3cm}
\begin{table}[htbp]
	\centering
	\caption{The change in optimal dimensionality of the word embeddings from basic to semantically infused corpus.}

	\label{tab:pip_loss}
	\begin{tabular}{l|c|c|c|c}
		\toprule
		& \multicolumn{1}{l}{} & \multicolumn{3}{c}{Algorithm}                                                         \\
		\midrule
		&                      & Word2Vec & GloVe & \begin{tabular}[c]{@{}c@{}}LSA\end{tabular} \\
		\midrule
		\multicolumn{1}{c}{\multirow{2}{*}{Corpus}} & Basic                & 29       & 30    & 26                                                                 \\
		\cmidrule{2-5}
		\multicolumn{1}{c}{}                        & Infused              & 29       & 28    & 27   \\
		\bottomrule                                                             
	\end{tabular}
\end{table}

\begin{figure}[htbp]
	\includegraphics[width=\linewidth]{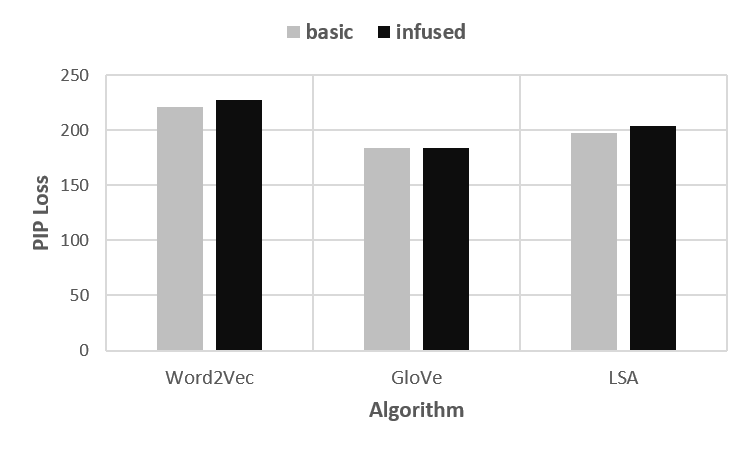}
	\vspace{-0.6cm}
	\caption{The change in PIP loss values corresponding to optimal dimensionality of the word embeddings from basic to semantically infused corpus.}
	
	\label{fig:pip_loss_comp}
\end{figure}

We compare the change in optimal dimensionality of the word embeddings (as shown in Table~\ref{tab:pip_loss}) and in the corresponding PIP loss values (as shown in Fig.~\ref{fig:pip_loss_comp}), from basic to the infused corpus, based on $3$ algorithms: Word2Vec~\cite{mikolov2013efficient}, GloVe~\cite{pennington2014glove} and Latent Semantic Analysis (LSA)~\cite{deerwester1990indexing}. We observe that there is an insignificant change in the optimal dimensionality and the corresponding PIP loss values from basic to the infused corpus. This demonstrates that the semantic infusion technique keeps the word relations and their quality intact while associating meta-data with the corpus text. Thus, this suggests that the semantic infusion technique is a near-lossless in nature.

\subsection{Filtration}
We leverage the anchored communities $C_A$ and Semantic Noise Cleansing ($\mathbf{N}$) module to filter out the semantically noisy sentences from the ticket text of the consumer complaints. We refer to the filtered sentences as semantic noise for the corresponding component class. The results after module $\mathbf{N}$ are summarized in the Table \ref{tab:sem_noi_rem}.

\subsection{Evaluation \& Results}

We evaluate the proposed framework in terms of how effective it is in removing the semantic noise of the problem component classes. We randomly select $100$ sentences for each of the $14$ problem component classes using univariate normal (Gaussian) distribution. An annotator (with moderate knowledge of the domain) manually tags the sentences as $1$ if the sentence is semantically noisy, else as $0$. Let $S_a$ represent the sentences tagged as semantic noise (by the annotator) for component class $a$ and $\hat{S}_a$ represent the sentences marked as semantic noise by the framework. Using these, we compute the performance measures: Precision, Recall and F1-score for each component class $a$ as follows:
\vspace{-0.1cm}
\begin{eqnarray}
Precision_a &=& \frac{|\hat{S}_a \cap S_a|}{|\hat{S}_a|} \\
Recall_a &=& \frac{|\hat{S}_a \cap S_a|}{|S_a|} \\
F1-Score_a &=& \frac{2*Precision_a*Recall_a}{Precision_a+Recall_a}
\end{eqnarray}

Table~\ref{tab:sem_noi_rem} presents the performance scores for the randomly selected sentences across $14$ problem component classes. We observe that the framework identifies the semantic noise with the highest precision of $0.97$ and an average precision of $0.81$. This indicates that the framework is effective in distinguishing between the semantic noise and meaningful information. Thus for various Industry setups, which usually demands high precision for corpus cleaning step in their decision-making process and activities, this framework can be a valuable asset.

%
\begin{table*}[h]
	\centering
	\caption{The statistics of NHTSA dataset after the Semantic Noise Cleansing ($\mathbf{N}$) module. Also, the evaluation results of the framework in terms of  Precision (P), Recall (R) \& F1-Score (F1) based on 100 Random Samples for each Component Class.}
	\vspace{0.1cm}
	\label{tab:sem_noi_rem}
	\begin{tabular}{l|c|c|c|c|c|c}
		\toprule
		Component Class      & \begin{tabular}[c]{@{}c@{}}\#\\ Sentences\end{tabular} & \begin{tabular}[c]{@{}c@{}}\#\\ Sentences \\Tagged as \\Semantic Noise\end{tabular} & \begin{tabular}[c]{@{}c@{}} Sentences\\ Tagged as\\ Semantic Noise\\ (in \%)\end{tabular} & P & R  & F1 \\
		\midrule
		Child-Seat                   & 11,671       & 4,004                                                                            & 34.31                                                                              & 0.84   & 0.51 & 0.64  \\
		\midrule
		Electronic-Stability-Control & 37,950       & 27,175                                                                           & 71.61                                                                             & 0.77   & 0.83 & 0.80  \\
		\midrule
		Equipment                    & 14,817       & 8,604                                                                            & 58.07                                                                              & 0.81   & 0.87 & 0.84  \\
		\midrule
		Exterior-Lighting            & 15,041       & 7,324                                                                            & 48.69                                                                               & 0.75   & 0.75 & 0.75  \\
		\midrule
		Fuel-Propulsion-System       & 33,716       & 23,926                                                                           & 70.96                                                                               & 0.97   & 0.76 & 0.85  \\
		\midrule
		Fuel-System                  & 9,355        & 4,345                                                                            & 46.45                                                                               & 0.73   & 0.72 & 0.72  \\
		\midrule
		Latches-Locks-Linkages       & 18,097       & 11,824                                                                           & 65.34\                                                                              & 0.70   & 0.64 & 0.67  \\
		\midrule
		Seat-Belts                   & 6,942        & 1,368                                                                            & 19.71                                                                              & 0.87   & 0.67 & 0.75  \\
		\midrule
		Seats                        & 12,844       & 5,019                                                                            & 39.08                                                                               & 0.94   & 0.82 & 0.88  \\
		\midrule
		Service-Brakes               & 5,868        & 773                                                                              & 13.17                                                                              & 0.69   & 0.56 & 0.62  \\
		\midrule
		Tires                        & 19,688       & 9,606                                                                            & 48.79                                                                               & 0.82   & 0.66 & 0.73  \\
		\midrule
		Visibility                   & 9,529        & 4,501                                                                            & 47.23                                                                              & 0.69   & 0.88 & 0.78  \\
		\midrule
		Visibility-Wiper             & 30,325       & 20,998                                                                           & 69.24                                                                              & 0.94   & 0.87 & 0.90  \\
		\midrule
		Wheels                       & 20,797       & 10,619                                                                           & 51.06                                                                               & 0.86   & 0.71 & 0.78 \\
		\bottomrule
	\end{tabular}
\end{table*}

\section{Conclusions \& Future Work}
\label{sec:concl}
We introduce a novel semantic infusion technique, which helps to have an association between the meta-data and text of a categorical corpus when represented in a vector space. We demonstrate the near-lossless nature of the technique using the PIP loss metric. To demonstrate the utility of this technique, we present a text-preprocessing framework to identify (in an unsupervised fashion) the semantic noise in a given categorical corpus. We evaluate the efficiency of the framework using a web forum dataset from the automobile domain. Further work should focus on various applications of the semantic infusion technique such as trends analysis using temporal meta-data, increasing explainability and interpretability of learned machine learning models. 


%
%
%

\begin{small}
\bibliographystyle{aaai}
\bibliography{reference}
\end{small}

\end{document}